\documentclass[twocolumn,final]{svjour3}       
\let\labelindent\relax         

\usepackage{times}
\usepackage{graphicx}\graphicspath{{fig/}} 
\usepackage{algorithm}
\usepackage{algorithmic}
\usepackage{multirow} 
\usepackage{multicol}
\usepackage[cmex10]{amsmath}
\usepackage{array}
\usepackage{mdwmath}
\usepackage{mdwtab}
\usepackage{rotating}
\usepackage{amssymb}
\usepackage{hyperref}
\usepackage{xspace}
\usepackage{mfirstuc}
\usepackage{marvosym}
\usepackage[list=off]{caption} 
\usepackage{units}
\def\makeheadbox{\relax}
\makeatletter\newcommand{\manuallabel}[2]{\def\@currentlabel{#2}\label{#1}}\makeatother

\usepackage{xcolor}

\colorlet   {lightorange}{orange!20}
\colorlet   {lightgrey}  {gray!20}

\usepackage{ulem} 
\colorlet{jb}{red}
\colorlet{mh}{red}

\usepackage[textwidth=2\marginparwidth,textsize=tiny,colorinlistoftodos]{todonotes}
\usepackage{marginnote}
\presetkeys{todonotes}{color=white,bordercolor=white,inline}{}

\colorlet{yz}{green}
\colorlet{jm}{cyan}
\colorlet{pr}{orange}
\colorlet{hl}{blue}
\colorlet{mwh}{teal}

\input{ext/tex/abbreviations}

\renewcommand  {\CCL    }{Constraint Consistent Learning\xspace}
\providecommand{\UPE    }{Unconstrained Policy Error\xspace}\providecommand{\NUPE   }{Normalised \UPE}
\providecommand{\CPE    }{Constrained Policy Error\xspace}  
\providecommand{\POE    }{Constrained Policy Error\xspace}  \providecommand{\NPOE   }{Normalised \POE}
\providecommand{\PPE    }{Constrained Policy Error\xspace}  \providecommand{\NPPE   }{Normalised \PPE}
\providecommand{\RBF    }{Radial Basis Function\xspace}
\providecommand{\API    }{Application Programmers Interface\xspace}
\providecommand{\GSL    }{GNU Scientific Library\xspace}
\usepackage{amssymb}
\usepackage{mathtools}

\mathchardef\mhyphen="2D   

\providecommand{\R}     {\mathbb{R}}          
\providecommand{\T}     {\top}                
\providecommand{\bO}    {\boldsymbol{0}}      
\providecommand{\I}     {\mathbf{I}}          

\providecommand{\estimated} [1]{\tilde{#1}}
\providecommand{\normalised}[1]{\hat{#1}}

\providecommand{\pinv}      [1]{{#1}^\dagger}

\providecommand{\nd}      {n}                              
\providecommand{\nk}      {k}                              
\providecommand{\Nd}      {\mathcal{\MakeUppercase{\nd}}}  
\providecommand{\Nk}      {\mathcal{\MakeUppercase{\nk}}}  

\providecommand{\bx}     {\mathbf{x}}         
\providecommand{\nx}     {p}                  
\providecommand{\dimx}   {\mathcal{\MakeUppercase{\nx}}} 

\providecommand{\btheta}  {\boldsymbol{\theta}}               

\providecommand{\bA}     {\mathbf{A}}         
\providecommand{\bb}     {\mathbf{b}}         
\providecommand{\nb}     {s}                  
\providecommand{\dimb}   {\mathcal{\MakeUppercase{\nb}}} 
\providecommand{\bN}     {\mathbf{N}}         
\providecommand{\bw}     {\mathbf{w}}         
\providecommand{\bv}     {\mathbf{v}}         
\providecommand{\bJ}     {\mathbf{J}}         

\providecommand{\bf}     {\mathbf{f}}         

\providecommand  {\bu}  {\mathbf{u}}          
\providecommand  {\nu}  {q}                   
\providecommand  {\dimu}{\mathcal{\MakeUppercase{\nu}}} 
\providecommand  {\bpi} {\boldsymbol{\pi}}    

\providecommand{\bxn}    {\bx_{\nd}}          
\providecommand{\bX}     {\mathbf{X}}         

\newcommand  {\bU}    {\MakeUppercase{\bu}} 
\newcommand  {\bV}    {\MakeUppercase{\bv}} 
\newcommand  {\bW}    {\MakeUppercase{\bw}} 

\renewcommand  {\nu}      {q}                   

\providecommand{\bwn}     {\bw_{\nd}}           
\providecommand{\bun}     {\bu_{\nd}}           
\providecommand{\bAn}     {\bA_{\nd}}           
\providecommand{\bbn}     {\bb_{\nd}}           
\providecommand{\bpin}    {\bpi_{\nd}}          
\providecommand{\bPn}     {\bP_{\nd}}           

\providecommand{\ebA}     {\estimated{\bA}}     
\providecommand{\ebN}     {\estimated{\bN}}     
\providecommand{\ebLambda}{\estimated{\bLambda}}
\providecommand{\ebpi}    {\estimated{\bpi}}    
\providecommand{\ebAn}    {\ebA_{\nd}}          
\providecommand{\ebNn}    {\ebN_{\nd}}          

\providecommand{\ebwk}    {\estimated{\bw}_\nk} 

\providecommand{\bPhi}    {\boldsymbol{\Phi}}   
\providecommand{\bLambda} {\boldsymbol{\Lambda}}
\providecommand{\dimPhii} {R}                   
\providecommand{\blambda} {\boldsymbol{\lambda}}

\renewcommand{\Nk}        {\mathcal{K}}         

\newcommand{\ba}         {\mathbf{a}}           
\newcommand{\nba}        {\normalised{\ba}}     

\newcommand{\bP}         {\mathbf{P}}

\newcommand{\bPi}         {\mathbf{\Pi}}
\newcommand{\bSigma}      {\mathbf{\Sigma}}
\newcommand{\bmu}         {\mathbf{\mu}}
\newcommand{\bomega}     {\mathbf{\omega}}
\providecommand{\Nss}      {\mathcal{\MakeUppercase{n_s}}}  

\providecommand{\bbeta} {\boldsymbol{\beta}}    
\providecommand{\dimbeta}{M}                    

\providecommand{\cdata}   {\lstinline|data|\xspace}    
\providecommand{\cmodel}  {\lstinline|model|\xspace}   
\providecommand{\cX}      {\lstinline|X|\xspace}       
\providecommand{\cU}      {\lstinline|U|\xspace}       
\providecommand{\cPi}     {\lstinline|Pi|\xspace}      
\providecommand{\cTs}     {\lstinline|Ts|\xspace}      
\providecommand{\cUn}     {\lstinline|Un|\xspace}      
\providecommand{\cM}     {\lstinline|M|\xspace}      

\providecommand{\cc}      {\lstinline|c|\xspace}      
\providecommand{\cs}      {\lstinline|s|\xspace}      
\providecommand{\cphi}    {\lstinline|phi|\xspace}    
\providecommand{\csearch} {\lstinline|search|\xspace}  
\providecommand{\ctheta}  {\lstinline|theta|\xspace}  
\providecommand{\coptimal}{\lstinline|optimal|\xspace} 
\providecommand{\cnmse}   {\lstinline|nmse|\xspace}   
\providecommand{\cfproj}  {\lstinline|f_proj|\xspace} 
\providecommand{\coptions}{\lstinline|options|\xspace} 
\providecommand{\cTolFun} {\lstinline|TolFun|\xspace} 
\providecommand{\cTolX}   {\lstinline|TolX|\xspace}   
\providecommand{\cMaxIter}{\lstinline|MaxIter|\xspace} 
\providecommand{\cstats}  {\lstinline|stats|\xspace}  
\providecommand{\cnmse}   {\lstinline|nmse|\xspace}   
   
\providecommand{\cvar}    {\lstinline|var|\xspace}    
\providecommand{\cdims}   {\lstinline|dim*|\xspace}    
\providecommand{\cdims}   {\lstinline|dim*|\xspace}

\providecommand{\ccllearnnhat  }{\lstinline|ccl_learna_nhat|}

\graphicspath{
{fig/}
}
\providecommand{\figurename}{Fig.}
\providecommand{\tablename}{Table}
\newcommand*{\sref}[1]{\S\ref{s:#1}}            
\newcommand*{\tref}[1]{\tablename~\ref{t:#1}}   
\newcommand*{\fref}[1]{\figurename~\ref{f:#1}}  
\newcommand*{\eref}[1]{(\ref{e:#1})}            
\newcommand*{\iref}[1]{\ref{i:#1}}              

\usepackage[inline]{enumitem}
\setlist{nolistsep}
\newcommand{\il}[1]{\begin{enumerate*}[label=(\roman*)]#1\end{enumerate*}}

\newcommand{\eg}{\textit{e.g.,}~} %
\newcommand{\ie}{\textit{i.e.,}~} %
\usepackage{listings}
\title{\LARGE\bf A Library for Constraint Consistent Learning}
\author{%
Yuchen Zhao$^1$ \and Jeevan Manavalan$^1$ \and Prabhakar Ray$^1$ \and Hsiu-Chin Lin$^2$ \and Matthew Howard$^1$%
}%
\date{}
\institute{\Letter 
            \hspace*{3ex}Jeevan Manavalan \at
            \hspace*{5ex}jeevan.manavalan@kcl.ac.uk     
           \and
          \newline
          $^1$ \hspace*{2.5ex}King's College London, Bush House, Strand Campus, 30,
          \hspace*{5ex}Aldwych, London WC2B 4BG - United Kingdom   
          \newline\newline
          $^2$   \noindent\hspace*{2.5ex}University of Edinburgh, 1.17. Bayes Centre,47 Potterrow, \hspace*{5.3ex}Edinburgh, EH8 9BT - United Kingdom
}

\begin{document}
\maketitle
\begin{abstract}
This paper introduces the first, open source software library for \CCL (CCL). It implements a family of data-driven methods that are capable of \il{\item learning state-independent and -dependent constraints, \item decomposing the behaviour of redundant systems into task- and null-space parts, and \item uncovering the underlying null space control policy}. It is a tool to analyse and decompose many everyday tasks, such as wiping, reaching and drawing. The library also includes several tutorials that demonstrate its use with both simulated and real world data in a systematic way. This paper documents the implementation of the library, tutorials and associated helper methods. The software is made freely available to the community, to enable code reuse and allow users to gain in-depth experience in statistical learning in this area.

\end{abstract}

\section{Introduction}\label{s:introduction} 

\noindent \newabbreviation{\CCL}{CCL} is a family of methods for learning different parts of the equations of motion of redundant and constrained systems in a data-driven fashion \cite{Howard2009b,Towell2010,Lin2015}. It is able to learn representations of self-imposed or environmental constraints \cite{Lin2015,LinHumanoid,Armesto2017}, decompose the movement of redundant systems into task- and null space parts \cite{LinTaskConstraint,Towell2010}, and uncover the underlying null space control policy \cite{Howard2009,Howard2008a}.
\CCL\ enables:
\begin{enumerate}
\item\label{i:enables-everyday} Learning the constraints encountered in many everyday tasks through various representations. 
\item\label{i:learn-pi} Learning the underlying control behaviour from movements observed under different constraints. 
\end{enumerate}

In contrast to many standard learning approaches that incorporate \emph{the whole of the observed motions into a single control policy estimate} \cite{schaal2003computational}, \CCL\ separates the problem into learning \il{\item the constraint representations and \item the underlying control policy}. This provides more flexibility to the robot in the reproduction of the behaviour, for example, in face of demonstration data of one behaviour recorded under different constraints, \CCL\ can learn a single policy that generalises across the constraints \cite{Howard2008a}.

\begin{figure}[t!] 
	\centering%
	\includegraphics[width=\linewidth]{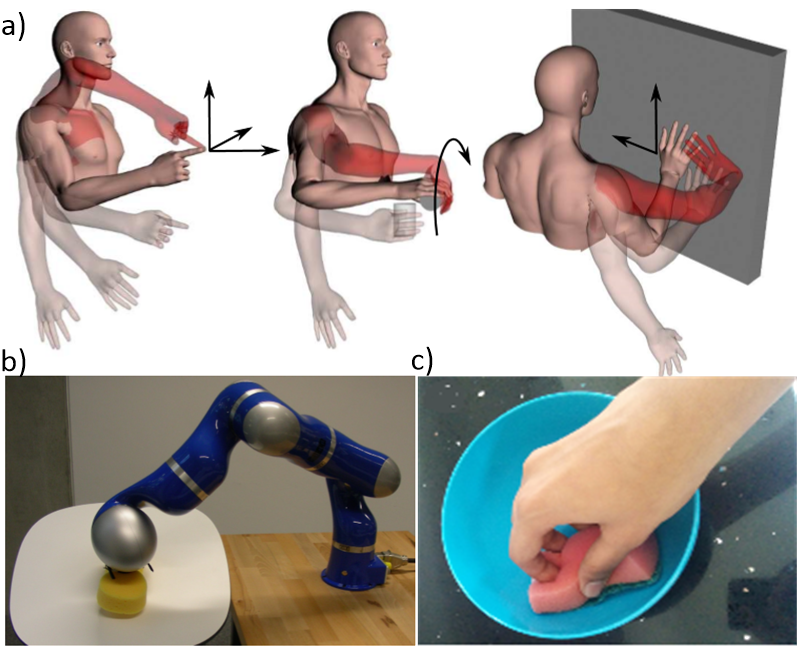}%
	\caption{\label{f:figure-label1}Typical examples in CCL. 
		\begin{enumerate*}[label=(\alph*)]
			\item\label{f:figure-label1a} Three different tasks: moving the finger to an $x,y,z$, position, liquid pouring and wiping a surface. In each case, redundancy is resolved in the same way. 
			\item\label{f:figure-label1b} A table wiping task is subject to a state-independent flat environment constraint using a robot arm.
			\item\label{f:figure-label1c} A bowl wiping task subject to a state-dependent curvature constraint.
		\end{enumerate*} 
}%
\end{figure}

In terms of \iref{enables-everyday}), the type of constraints may fall into the categories of \emph{state independent} or \emph{state dependent} constraints. For example, for state-space represented as end-effector coordinates, when wiping a table (see \fref{figure-label1}\ref{f:figure-label1b}), the flat surface acts as a hard restriction on the actions available (motions perpendicular to the surface will be eliminated by the constraint), regardless of where the effector is located on the surface, so represents a state independent constraint. When wiping or stirring soup in a bowl (see \fref{figure-label1}\ref{f:figure-label1c}), the curved surface introduces a state dependency in the constraint, since the restriction of motion is dependent on the location of the effector. 
The ability to predict how the constraints can influence the outcomes of control actions can speed-up learning of the new skills and enhance safety (\eg the learned constraints can prevent exploration actions that cause excessive force). Furthermore, the availability of constraint knowledge can reduce the dimension of the search space when optimising behaviours \cite{Bitzer2010}.

In terms of \iref{learn-pi}), the observable movements may contain partial information about the control policy masked by the constraint, or higher priory task objectives \cite{Towell2010}. For example, in a reaching task (see \fref{figure-label1}\ref{f:figure-label1a}), humans move their arms towards the target (primary task) while minimising effort, for instance, by keeping the elbow low (secondary objective). The imperative of executing the primary task represents a \emph{self-imposed constraint}, that restricts the secondary objective.  Interaction with \emph{environmental constraints}, can also mask the actions applied by the demonstrator. For example, when grasping an object on a table such as a pen, the fingers slide along the table surface resulting in a discrepancy between the observed motion and that predicted by the applied actions were the surface different in shape or orientation.
\CCL\ can help uncover these components, enabling the intended actions to be reconstructed and applied to new situations \cite{LinEtAl2018}.

\begin{figure*}[t!] 
	\includegraphics[width=\textwidth, height=6cm]{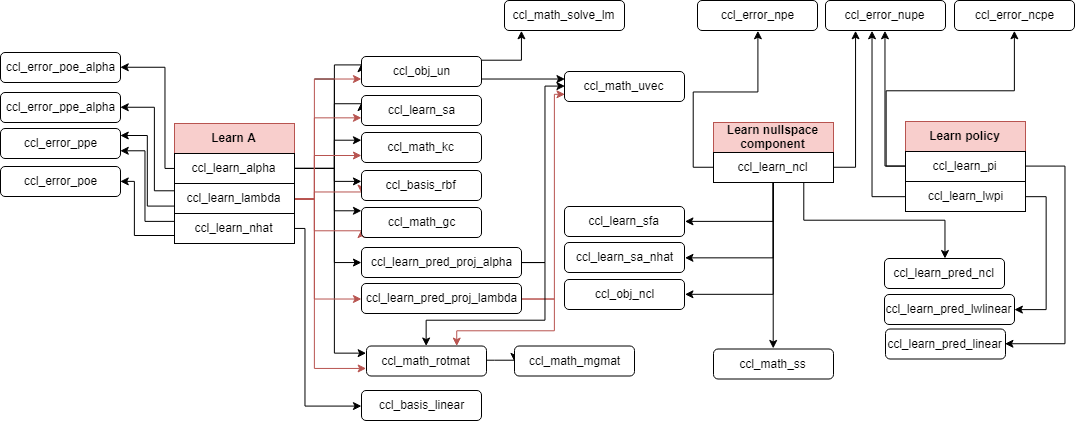}%
	\caption{A system diagram of the \CCL library. The complete dependency tree can be found in the package documentation. The library has been written under three categories: \texttt{learn A}, \texttt{learn null-space component} and \texttt{learn policy}. The arrows links the dependencies of the implementations to the main functions.} 
	\label{f:function_diagram}
\end{figure*}

This paper introduces the \emph{\CCL\ library} as a community resource for learning in the face of kinematic constraints, alongside tutorials to guide users who are not familiar with learning constraint representations and underlying policies. As its key feature, \CCL\ is capable of decomposing many everyday tasks (\eg wiping \cite{Lin2015}, reaching \cite{Towell2010} and gait \cite{Lin2014a}) and learn subcomponents (\eg null space components \cite{Towell2010}, constraints \cite{Lin2015} and unconstrained control policy \cite{Howard2008a}) individually. The methods implemented in the library provide means by which these quantities can be extracted from the data, based on a collection of algorithms developed since \CCL\ was first conceived in 2007 \cite{Howard2007}. It presents the \emph{first unified \emph{\newabbreviation{\API}{API}} implementing and enabling use of these algorithms in a common framework}. 



\section{Application Domain}\label{s:problem-definition}
\noindent The \CCL\ library addresses the problem of learning from motion subject to various constraints for redundant systems. The behaviour of constrained and redundant systems is well-understood from an analytical perspective \cite{Udwadia1996}, and a number of closely related approaches have been developed for managing redundancy at the levels of dynamics \cite{Peters2008a}, kinematics \cite{Liegeois1977,Whitney1969}, and actuation \cite{Howard2011,Tahara2009}. However, a common factor in these approaches is the assumption of prior knowledge of the constraints, the control policies, or both. \CCL\ addresses the case where there are uncertainties in these quantities and provides methods for their estimation.



\subsection{Constraint Formalism}
\noindent The \CCL\ library assumes training data to come from systems subject to the following formalism.

The system is considered to be subject to a set of $\dimb$-dimensional ($\dimb\leq\dimu$) constraints with the general form
\begin{equation}
	\bA(\bx)\bu(\bx) = \bb(\bx)
	\label{e:equation-1}
\end{equation}
where $\bx\in\R^\dimx$ represents state and $\bu\in\R^\dimu$ represents the action. $\bA(\bx)\in\R^{\dimb\times\dimu}$ is the \textit{constraint matrix} which projects the task space policy onto the relevant part of the control space. The vector term $\bb\in\R^\dimb$, if present, represents constraint-imposed motion (in redundant systems, it is the policy in task space). 

Inverting \eref{equation-1}, results in the relation
\begin{equation}
	\bu(\bx)=\underbrace{\pinv{\bA(\bx)}\bb(\bx)}_{\bv}+\underbrace{\bN(\bx)\bpi(\bx)}_{\bw}
	\label{e:equation-2}
\end{equation}
where $\pinv{\bA}$ denotes the unique Moore-Penrose pseudo-inverse of the matrix $\bA$, $\bN(\bx)=(\I-\pinv{\bA(\bx)}\bA(\bx))\in\R^{\dimu\times\dimu}$ is a projection matrix that projects the policy $\bpi(\bx)$ onto the null space of the constraint. Note that, the constraint can be state dependent ($\bA(\bx)$) or state independent ($\bA$). $\bv(\bx)\equiv\pinv{\bA(\bx)}\bb(\bx) $ and $\bw(\bx)\equiv\bN(\bx)\bpi(\bx)$ are termed the task-space and the null-space component, respectively.
Typically, it is assumed that the only directly observable quantities are the state-action pairs $(\bxn,\bun)$ $\nd\in1,\cdots,\Nd$ which contain an unknown combination of $\bv(\bx)$ and $\bw(\bx)$, or at greater granularity $\bA(\bx)$, $\bb(\bx)$, $\bN(\bx)$ and $\bpi(\bx)$. 
The \CCL\ library provides methods for estimating each of these quantities---implementation details are provided in \sref{method}. 

Note that, application of learning approaches \cite{klanke2008library} that do not consider the composition of the data in terms of the constraints are prone to poor performance and modelling errors. For example, applying direct regression to learn policies where there are variations in the constraints can result in model averaging effects that risk unstable behaviour \cite{Howard2009b}. This is due to factors such as \il{\item the \textit{non-convexity} of observations under different constraints, and \item \textit{degeneracy} in the set of possible policies that could have produced the movement under the constraint}. Providing an open-source collection of software tools suitable for application to this class of learning problems, can help those working in the field to avoid these pitfalls.

\section{Implementation}\label{s:method}
\noindent A system diagram of the \CCL\ library is shown in \fref{function_diagram}. The naming convention is following \lstinline|ccl_xxx_xxx| indicating the category and functionality of the implementations. The following provides brief notes on the implementation languages and library installation, explains the data structures used in toolbox.
\subsection{Language and Installation Notes}
\noindent The library is implemented in both \textit{C} and \textit{Matlab} and is available for download from Github \footnote{\url{www.github.com/mhoward3210/ccl}}. 
The \CCL\ library is provided under the \textit{GNU General Public License v3.0}, 
and documentation is provided online\footnote{\url{nms.kcl.ac.uk/rll/CCL\textunderscore doc/index.html}}.
To use the Matlab package, the user can just add the \CCL\ library into the current path. Installation of the C package has been tested on Ubuntu 14.04 Linux systems and is made easy through use of the \texttt{autoconf} utility. For Windows systems, only the Matlab package is currently tested and supported.
The C package's only dependency is on the 3rd-party, open-source \newabbreviation{\GSL}{GSL}\footnote{\url{www.gnu.org/software/gsl/}}.

A detailed explanation of the functions follows, alongside tutorial examples that are provided to enable the user to easily extend use of the library to their chosen application (see \sref{tutorials}). 


\subsection{Data Structures}\label{s:method_data_structure}
\noindent The methods implemented in the toolbox work on data that is given as tuples $\{\bxn,\bun\}_{\nd=1}^\Nd$ of observed states and constrained actions. It is assumed that all the actions $\bu$ are generated using the same underlying policy $\bpi(\bx)$. In particular\footnote{For brevity, here, and throughout the paper, the notation $\mathbf{a}_\nd$ is used to denote the $\nd$th sample of the (matrix or vector) quantity $\textbf{a}$. Where that quantity is a function of $\bx$, the notation $\mathbf{a}_\nd$ denotes the quantity calculated on the $\nd$th sample of $\bx$, \ie $\mathbf{a}_\nd=\mathbf{a}(\bxn)$.} $\bun=\pinv{\bAn}\bbn+\bN_{\nd}\bpin$, where $\bAn$, $\bbn$ and $\bpin$ are not explicitly known. The observations are assumed to be grouped into $\Nk$ subsets of $\Nd$ data points, each (potentially) recorded under different (task or environmental) constraints. 

The \CCL\ library uniformly stores the data tuples in a data structure reflecting this problem structure. The latter includes the data fields incorporating samples of the input state ($\bx$), action ($\bu$), actions from unconstrained policy ($\bpi$), task space component ($\bv$) and null space component ($\bw$). The major data structure and data fields are listed in \tref{data_struct}.

\begin{table}[]
\centering
\caption{Main data structures of the \CCL\ library. \textnormal{\cdims} indicates the data fields for definition of the data dimensionality.}
\label{t:data_struct}
\scalebox{.67}{%
\begin{tabular}{cccc}
\hline
		Struct    & Main Fields & Definition                                            & Other Fields \\\hline
		\cdata    & \cX         & $\bX =(\bx_1, \cdots,\bx_\Nd )\in\R^{\dimx\times\Nd}$ &              \\
        ~         & \cU         & $\bU =(\bu_1, \cdots,\bu_\Nd )\in\R^{\dimu\times\Nd}$ &              \\
        ~         & \cPi        & $\bPi=(\bpi_1,\cdots,\bpi_\Nd)\in\R^{\dimu\times\Nd}$ &              \\
        ~         & \cTs        & $\bV =(\bv_1, \cdots,\bv_\Nd )\in\R^{\dimu\times\Nd}$ &              \\
        ~         & \cUn        & $\bW =(\bw_1, \cdots,\bw_\Nd )\in\R^{\dimu\times\Nd}$ &              \\\hline
		\cmodel   & \cM      & Nullspace component Model weighting parameters $\bomega$.                 &              \\
        ~         & \cc         & $\bmu$                                                &              \\
        ~         & \cs         & $\bSigma$                                             &              \\
        ~         & \cphi       & $\bbeta$                                              &              \\
        ~         &             &                                                       & \cdims       \\\hline
		\csearch  & \ctheta     & $(\btheta_1,\cdots,\btheta_{\Nss})$                   &              \\
        ~         &             &                                                       & \cdims       \\\hline
		\coptimal & \cnmse      & Normalised mean square error.                         &              \\
        ~         & \cfproj     & Function handle for projection matrix.                &              \\
        ~         &             &                                                       & \cdims       \\\hline
		\coptions & \cTolFun    & Tolerance for residual.                               &              \\
        ~         & \cTolX      & Tolerance for model parameters.                       &              \\
        ~         & \cMaxIter   & Maximum iterations for optimiser.                     &              \\
        ~         &             &                                                       & \cdims       \\\hline
		\cstats   & \cnmse      & Normalised mean square error.                         &              \\
        ~         & mse      & Mean squared error.                                   &              \\
        ~         & \cvar       & Variance.                                             &              \\\hline
\end{tabular}
}
\end{table}
Learning is more effective when the data contains sufficiently rich variations in one or more of the quantities defined in \eref{equation-2}, since methods learn the consistency by teasing out the variations. For instance, when learning $\bv$, variations in $\bw$ are desirable \cite{Towell2010}. For learning $\bpi$, observations subject to multiple constraints (variation in $\bA$) are necessary \cite{Howard2009b}. For learning constraint $\bA$, variations in $\bpi$ are desirable \cite{LinTaskConstraint}.

\section{Learning Functions}\label{s:learning_functions}
The \API of the \CCL library provides functions for a series of algorithms for the estimation of the quantities $\bA$, $\bv$, $\bw$ and $\bpi$. The following provides a summary of the methods provided, including implementation notes and a brief summary\footnote{Due to space constraints, the reader is referred to the original research literature for in-depth details of the theoretical aspects.} of the theoretical basis for each algorithm.

\subsection{Learning the Constraint Matrix}\label{s:method_learn_constraint}
\noindent The library implements several methods for estimating the constraint matrix $\bA$. 
These assume that \il{%
\item each observation contains a constraint $\bA\neq\bO$,
\item the observed actions take the form $\bu=\bw=\bN\bpi$,
\item $\bu$ are generated using the same null space policy $\bpi$, and 
\item $\bA$ is not explicitly known for any observation
} (optionally, features in form of candidate rows of $\bA$, may be provided as prior knowledge--see \sref{learn-state-dependent-A}).

{\renewcommand{\bu}{\bw}
The methods for learning $\bA$ under these assumptions, are based the insight that \textit{the projection of $\bu$ also lies in the same image space} \cite{Lin2015}, \ie
\begin{equation}
	(\I-\pinv{\bA}\bA)\bu=\bN\bu=\bu.	
	\label{e:Nw=w}
\end{equation}
\noindent If the estimated constraint matrix $\ebA$ is accurate, it should also obey this equality. Hence, the estimate can be formed by minimising the difference between the left and right hand sides of \eref{Nw=w}, as captured by the error function
\begin{equation}
	E[\ebN]=\sum^{\Nd}_{\nd=1} ||\bun-\ebNn\bun||^{2}
    \label{e:equation-4}
\end{equation}
\noindent where $\ebNn=\I-\pinv{\ebAn}\ebAn$. This can be expressed in simplified form
\begin{equation}
	E[\ebA]=\sum^{\Nd}_{\nd=1} ||\bun^\T\pinv{\ebAn}\ebAn\bun||^2.
	\label{e:equation-5}
\end{equation}
Note that, in forming the estimate, \emph{no prior knowledge of $\bpi$ is required}. However, an appropriate representation of the constraint matrix is needed.\footnote{Note: While the pseudoinverse operation has conditioning problems when close to a rank-deficient configuration, the constraint learning methods deal with this during the inversion by ignoring all singular values below a certain threshold.}  As outlined in \sref{introduction}, in the context of learning $\bA$, there are two distinct cases depending on whether the constraints are state-dependent or not. The \CCL\ library implements methods for both scenarios.


\subsubsection{Learning State Independent Constraints}\label{s:learn-state-independent-A}
The simplest situation occurs if the constraint is unvarying throughout the state space. In this case, the constraint matrix can be represented as the concatenation of a set of unit vectors 
\begin{equation}
	\ebA=(\nba_1^\T, \nba_2^\T, \cdots, \nba_{\dimb}^\T)^\T
	\label{e:equation-6}
\end{equation}
where the $\nb$th unit vector $\nba_\nb$ is constructed from the parameter vector $\btheta\in\R^{\dimu-1}$ that represents the orientation of the constraint in action space (see \cite{Lin2015} for details).
\begin{figure}[t!] 
\centering
\includegraphics[width=.48\textwidth]{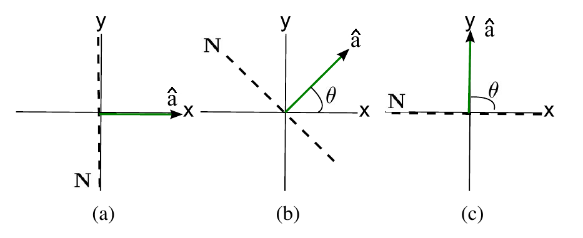}
\caption{\label{f:thetaanda} Three one-dimensional unit vector constraints $\nba$ where 
\begin{enumerate*}[label=(\alph*)]
			\item\label{f:thetaandaa} $\theta = 0^\circ$, 
			\item\label{f:thetaandab} $\theta = 45^\circ$, and
			\item\label{f:thetaandac} $\theta = 90^\circ$
		\end{enumerate*} 
(reproduced from \cite{Lin2015}).}
\end{figure}
A visualisation of the simplest case of a one-dimensional constraint in a system with two degrees of freedom where $\bA=\nba\in\R^{1\times2}$ is shown in \fref{thetaanda}. It can be seen how $\btheta$ (in this case, $\nba=(\cos{\theta},\sin{\theta})$) corresponds to different estimates of the constraint its associated null-space.

The \CCL library provides the function 
$
\begin{minipage}{3.1764in}\vspace{2ex}
\scalebox{1}{\coptimal\lstinline|=|\ccllearnnhat\lstinline|(|\cUn\lstinline|)|}
\end{minipage} 
 $
\newline \newline
\noindent to form the estimate $\ebA$ in this scenario, where \cUn is the collated set of samples of $\bu$ (see \tref{data_struct}) and \coptimal is the learnt model. The C counterpart\footnote{The C library implements all learning functions, but for the remainder of the paper, Matlab functions will be used throughout for consistency.} of this function is
{\lstset{basicstyle=\footnotesize\ttfamily}
$\hspace{0em}%
\begin{minipage}{3.1764in}
\begin{lstlisting}
void learn_nhat
     (const double *Un,const int dim_u,
	   const int dim_n,NHAT_Model *optimal)
\end{lstlisting}
\end{minipage}
$
}
\noindent where the input argument \cUn and output argument \coptimal in the C library is identical with Matlab implementation. The arguments \lstinline|dim_u| and \lstinline|dim_n| are used for defining the dimensionality of the array \cUn.

The function works by searching for rows of the estimated constraint matrix $\ebA$ using the function
$
\begin{minipage}{2.1764in}
\begin{lstlisting}
[model, stats] = 
   ccl_learna_sfa (UnUn,Un,model,search)
\end{lstlisting}
\end{minipage}
$

\noindent the arguments \cmodel and \csearch contain the initial model (untrained) model, parameters controlling the search, respectively  
and \lstinline|UnUn| 
is the matrix $\bW^\T\bW$ where $\bW$ is a matrix containing samples of $\bw$. This is pre-calculated prior to the function call for speed. The function returns the learnt model \cmodel and learning statistics \cstats (see \tref{data_struct}).

}
\subsubsection{Learning State Dependent Constraint}\label{s:learn-state-dependent-A}
Another proposed method for learning a state-dependent constraint, \textit{i.e.}, the constraint $\bA(\bx)$ depends on the current state of the robot $\bx$. For learning $\bA(\bx)$, two scenarios may be encountered where \il{\item there is prior knowledge of the constraint in form of candidate rows (see \cite{LinTaskConstraint}), or \item no prior knowledge is available}.

In the case of (i),
\begin{equation}
	\bA(\bx)=\tilde{\bLambda}(\bx)\bPhi(\bx).
	\label{e:equation-10}
\end{equation}
$\bPhi(\bx)\in\R^{\dimPhii\times\dimu}$ is the feature matrix and $\tilde{\bLambda}(\bx)\in\R^{\dimb\times\dimPhii}$ is the \textit{selection matrix} specifying which rows represent constraints. The feature matrix can take the form $\bPhi(\bx)=\bJ(\bx)$, where $\bJ$ is the Jacobian mapping from the joint space to the end-effector task space. Similar to \eref{equation-6}, $\ebLambda(\bx)$ is described by a set of $\dimb$ orthonormal vectors
\begin{equation}
	\ebLambda(\bx) = (\blambda _1(\btheta)^\T,\blambda _2(\btheta)^\T,\cdots,\blambda _{\dimb}(\btheta)^\T )^\T
	\label{e:equation-11}
\end{equation}
\noindent where $\blambda_s\in\R^\dimb$ is the $s$th dimension in the task space and $\blambda_{i} \perp \blambda_{j}$ for $i\neq j$. Parameter $\btheta_s\in\R^{\dimb-1}$ is used for representing the constraint vector $\blambda$. Each $\btheta_s$ is modelled as $\btheta_s=\bomega_s\bbeta(\bx)$ where $\bomega_s\in\R^{(\dimu-1)\times G}$ is the weight matrix, and $\bbeta(\bx)\in\R^G$ is the vector of $G$ basis functions that transform $\bx$ into a set of feature vectors in higher dimensional space. Substituting $\ebA_{\nd}=\ebLambda_{\nd}\bPhi_{\nd}$, \eref{equation-5} can be written as
 \begin{equation}
 E[\ebLambda]=\sum^{\Nd}_{\nd=1}||\bun^\T\pinv{(\ebLambda_{\nd}\bPhi_{\nd})} \ebLambda_{\nd}\bPhi_{\nd}\bun||^2.
	\label{e:equation-12}
\end{equation}
The optimal $\ebLambda$ can be formed by iteratively searching the choice of $\blambda_s$ that minimises \eref{equation-12}.

The objective function for \eref{equation-5} and (\eref{equation-12}) are embedded in function

\begin{minipage}{-0.23in}
\begin{lstlisting}[xleftmargin=1\textwidth]
[fun] = ccl_obj_AVn (model,W,BX,RnUnUnn)
\end{lstlisting}
\end{minipage}

\noindent where \texttt{fun} is the objective function handle, \texttt{M} is the model parameter $\bomega$, \texttt{BX} is the feature vectors in high dimension space $\bbeta(\bx)$, and \texttt{RnUnUnn} is the pre-rotated $UnUn_n$.

The implementation of case (i) is:

\hspace{-0.8cm}\begin{minipage}{3in}
\begin{lstlisting}
 [optimal] = 
    ccl_learn_lambda (Un,X,Phi,options)
\end{lstlisting}
\end{minipage}

\noindent Where \texttt{Phi} is the feature matrix $\bPhi$.
The implementation of case (ii) where 
\begin{equation}
	\bA(\bx) = (\ba_1(\btheta)^\T,\ba _2(\btheta)^\T,\cdots,\ba_{\dimb}(\btheta)^\T )^\T
    \label{e:equation-11case2}
\end{equation}
is:

\hspace{-0.8cm}\begin{minipage}{3.30in}
\begin{lstlisting}
 [optimal] = 
    ccl_learn_alpha (Un,X,options)
\end{lstlisting}
\end{minipage}

\noindent The non-linear parameter optimisation solver used for learning $\btheta_s$ calculates the direction in which the parameter is improved for this, the Levenberg-Marquardt (LM) solver is used. The latter is embedded in the function



\hspace{-0.65cm}\begin{minipage}{2.1764in}
\begin{lstlisting}
[xf,S,msg] = ccl_math_solve_lm (varargin)
\end{lstlisting}
\end{minipage}

\noindent where the inputs \texttt{varargin} are the objective function (\texttt{fun}), initial guess of model parameters (\texttt{xc}, \ie$\bomega$) and training options (\texttt{options}).

\subsection{Learning the Task and Null Space Components}\label{s:method_B}
\noindent When the unconstrained control policy is subject to both constraint and some task space policy ($\bb(\bx)\neq\bO$), it is often useful to extract the task and null space components ($\bv(\bx)$ and $\bw(\bx)$, respectively) of the observed actions $\bu(\bx)$. The \CCL\ library provides the following methods to estimate these quantities. These assume that the underlying null-space policy and the task constraint are consistent.



Assuming that the underlying null-space policy $\bpi(\bx)$ and the task constraint $\bA(\bx)$ are consistent across the data set, the null space component $\bw(\bx)$ should satisfy the condition $\bN(\bx)\bu(\bx)=\bw(\bx)$ as noted in \cite{Towell2010}, in this case, $\bw(\bx)$ by minimising
\begin{equation}
	E[\tilde{\bw}] = \sum^{\Nd}_{\nd=1}||\tilde{\bP}_n \bu_n-\tilde{\bw}(\bxn)||^2
	\label{e:equation-20}
\end{equation}

\noindent with $\tilde{\bP}_n = \tilde{\bw}_n \tilde{\bw}_n^\T /||\tilde{\bw}_n||^2$, where $\tilde{\bw}_n=\tilde{\bw}(\bxn)$ projects an arbitrary vector onto the same space of $\bw$, and  $\bun$ is the $\nd$th data point. This error function eliminates the task-space components $\bv(\bx)$ by penalising models that are inconsistent with the constraints, \ie those where the difference between the model, $\tilde{\bw}(\bx)$, and the observations \textit{projected onto that model} is large (for details, see \cite{Towell2010}). In the \CCL\ library, this functionality is implemented in the function

\hspace{-0.6cm}\begin{minipage}{2.1764in}
\begin{lstlisting}
[fun,J] = ccl_obj_ncl (model,W,BX,U)
\end{lstlisting}
\end{minipage}

\noindent where \texttt{U} is the observed actions $\bU$. The outputs are \texttt{fun} (function handle) and \texttt{J} (analytical Jacobian of the objective function \eref{equation-20}).

Minimisation of \eref{equation-20} penalises models that are inconsistent with the constraints, \ie those where the difference between the model, $\ebwk(\bx)$, and the observations \textit{projected onto that model} is large (for details, see \cite{Towell2010}). In the \CCL\ library, this functionality is implemented in the function

\hspace{-0.9cm}\begin{minipage}{2.1764in}
\begin{lstlisting}
  model = ccl_learnv_ncl(X,U,model)
\end{lstlisting}
\end{minipage}

\noindent where the inputs are \texttt{X} (input state), \texttt{Y} (observed actions combined with task and null space components) and \cmodel (model parameters). The output is the learnt model \cmodel.

\subsection{Learning the Unconstrained Control Policy}

\noindent The \CCL\ library also implements methods for estimating the underlying \emph{unconstrained control policy} ($\bpi(\bx)$). These assume either (i) $\bb=0$ no additional task is involved or (ii) $\bw$ is learnt using the method proposed in \sref{method_B}.

As shown in \cite{Howard2008a}, an estimate $\ebpi(\bx)$ can be obtained by minimising the \emph{inconsistency error}
\begin{equation}
	E_\pi = \sum^{\Nd}_{\nd=1} ||\bun-\bPn\ebpi(\bxn)||^2;\quad \bPn=\frac{\bun\bun^\T}{||\bun||^2}.
	\label{e:equation-17}
\end{equation}

\noindent The risk function \eref{equation-17} is compatible with many regression models. The \CCL\ library implements a parametric policy learning scheme. Where $\ebpi(\bx)=\bW\bbeta(\bx)$, $\bW\in \R^{\dimu\times\dimbeta}$ is a matrix of weights and $\bbeta(\bx)\in \R^\dimbeta$ is a vector of fixed basis functions (\eg linear features or Gaussian radial basis functions). A locally-weighted linear policy learning is also implemented in the library to improve the robustness of the policy learning (Details see \cite{Towell2010}). A reinforcement learning scheme (with possibly deep network structure) can also be used to learn more complex policies \cite{Howard2013,Howard2013b,Gu2016}.

The \CCL\ library implements the learning of these models through the functions

\hspace{-0.9cm}\begin{minipage}{2.1764in}
\begin{lstlisting}
  model = ccl_learnp_pi (X,U,model)
\end{lstlisting}
\end{minipage}


\noindent where inputs \texttt{X} and \texttt{U} are observed states and actions. The outputs are the learnt model \cmodel.

\section{Evaluation Criteria}\label{s:evaluation_criteria}
\noindent For testing different aspects of the learning performance, a number of evaluation criteria have been defined in the \CCL\ literature. These include metrics comparing estimation against the ground truth (if known), as well as those that provide a pragmatic estimate of performance based only on observable quantities. In the context of \textit{evaluating the constraints}, the \CCL\ library provides implementations of the following functions.

If the ground truth constraint matrix $\bA_\nd$ and unconstrained policy $\bpi_{\nd}$ are known, then the \textit{normalised projected policy error} (\NPPE) provides the best estimate of learning performance \cite{Lin2015}. In the \CCL\ library, The \NPPE is computed through the functions

\hspace{-0.6cm}\begin{minipage}{2.1764in}
\begin{lstlisting}
[nPPE,vPPE,uPPE] = 
    ccl_error_ppe (U_t,N_p,Pi)
\end{lstlisting}
\end{minipage}

\noindent where \texttt{U\_t} is the true null space components $\bW$, \texttt{N\_p} is the learned projection matrix $\bN$, and \texttt{Pi} is the unconstrained control policy $\bpi$. The outputs are the \NPPE, variance and (non-normalised) mean-squared PPE, respectively.

In the absence of ground truth $\bA_{\nd}$ and $\bpi_{\nd}$, the \textit{normalised projected observation error} (\NPOE) must to be used \cite{Lin2015}. The functions implemented for computing the \NPOE are

\hspace{-0.6cm}\begin{minipage}{2.1764in}
\begin{lstlisting}
[nPOE,vPOE,uPOE] = 
    ccl_error_poe (U_t,N_p,Pi)
\end{lstlisting}
\end{minipage}
\newline
\noindent where the inputs and outputs conventions are similar to those of the functions computing the \NPPE.

To evaluate the predictions of the \textit{null-space component model} $\ebwk(\bx)$, the \textit{null space projection error} (NPE) can be used \cite{Towell2010}. This is  implemented in the function 

\hspace{-1cm}\begin{minipage}{2.1764in}
\begin{lstlisting}
  [umse,v,nmse] = ccl_error_npe (Un,Unp)
\end{lstlisting}
\end{minipage}
\newline
\noindent where \texttt{Un} and \texttt{Unp} are the true and predicted null space component. The return values are the NPE, variance and (non-normalised) mean squared projection error. Note that, use of this function assumes knowledge of the ground truth $\bwn$. 

\begin{table*}[t]
\centering
\caption{Key Functions in the CCL Library}
\label{t:function_overview}
\scalebox{1.20}{
\begin{tabular}{ccc}
\hline
Function Type                     & Name        & Description                            \\ \hline
\multirow{3}{*}{Learn Constraint} & Learn Alpha & Learning state dependent projection N \eref{equation-12} + \eref{equation-11}\\
                                  & Learn Lambda           & Learning state dependent selection matrix $\ebLambda$ \eref{equation-12}  + \eref{equation-11case2}                                     \\
                                  & Learn Nhat           & Learning state independent constraint N  \eref{equation-5}                                    \\ \hline
Learn Nullspace Component         & Learn ncl           & Learn the nullspace component  of u = v+w \eref{equation-20}                    \\ \hline
Learn unconstrained control Policy    & Learn CCL Pi  & Learn null space policy   \eref{equation-17}               \\ \hline
\multirow{2}{*}{Evaluate Constraint error}              & NPPE           & \begin{tabular}[c]{@{}l@{}}normalised projected policy error\end{tabular}     \\
                                  &       NPOE      &  \begin{tabular}[c]{@{}l@{}} normalised projected observation error\end{tabular}     \\ \hline
Evaluate Nullspace component error  &   NPE          &   nullspace projection error                                       \\ \hline
\multirow{2}{*}{Evaluate unconstrained control policy }            &   NUPE          &         normalised unconstrained policy error     \\
  &      NCPE       &    normalised constrained policy error \\ \hline
\end{tabular}
}
\end{table*}

To evaluate the \textit{estimated unconstrained control policy model} $\ebpi(\bx)$, the \textit{normalised unconstrained policy error} (NUPE) and \textit{normalised constrained policy error} (NCPE) \cite{Howard2008a} are used. The former assumes access to the ground truth unconstrained policy $\bpi_{\nd}$, while the latter assumes the true $\bwn$ is known. They are implemented in the functions 

\hspace{-1cm}\begin{minipage}{2.1764in}
\begin{lstlisting}
  [umse,v,nmse] = ccl_error_nupe (F,Fp)
\end{lstlisting}
\end{minipage}

\noindent and

\hspace{-1cm}\begin{minipage}{2.1764in}
\begin{lstlisting}
  [umse,v,nmse] = ccl_error_ncpe(F,Fp,P)
\end{lstlisting}
\end{minipage}

\noindent where input \texttt{F} is the true unconstrained control policy commands, \texttt{Fp} is the learned unconstrained control policy and \texttt{P} is the projection matrix. The outputs are \NUPE (respectively, NCPE), the sample variance, and the mean squared \UPE (respectively, \CPE).

\tref{function_overview} shows an overview of all key functions as well as their respective equations for learning and evaluation which are covered in this section.

\section{Tutorials}\label{s:tutorials}
\noindent The \CCL\ library package provides a number of tutorial scripts to illustrate its use. Multiple examples are provided which demonstrate learning using both simulated and real world data, including, learning in a simulated 2-link arm reaching task, and learning a wiping task. The following describes the toy example in detail, alongside a brief description of other included demonstrations which were tested with real world systems. Further details of the latter are included in the documentation.

In the first, a simple, illustrative toy example is used, in which a two-dimensional system is subject to a one-dimensional constraint. In this, the learning of \il{\item the null space components, \item the constraints, and \item the null space policy} is demonstrated.  

The toy example code has been split into three main sections for learning different parts of the equations. Detailed comments of the functions can be found in the Matlab script. The procedure for generating data is dependant on the part of the equation you wish to learn. Details can be found in the documentations.

\subsubsection{Learning Null Space Components $\bw$}

\noindent This section gives a guidance of how to use the library for learning $\bW$. A sample Matlab snippet is shown in \tref{code_learn_ns} and explained as follows. Firstly, the user needs to generate training data given the assumption that $\bw$ is fixed but $\bv$ is varying in your demonstration. The unconstrained policy is a limit cycle policy \cite{LinHumanoid}. For learning, the centre of the \RBF s are chosen according to K-means and the variance as the mean distance between the centres. A parametric model with 16 Gaussian basis functions is used. Then the null-space component model can be learnt through \texttt{\textit{ccl\_learnv\_ncl}}. Finally, the evaluation metrics \texttt{\textit{NUPE}} and \texttt{\textit{NPE}} are used to report the learning performance.

\subsubsection{Learning Null Space Constraints $\bA$}
This section gives guidance of how to use the library to learn a state independent $\bA$ and state dependent $\bA(\bx)$, where $\bb=0$. A sample Matlab script is provided in \tref{code_learn_A} and explained as following. Firstly, it simulates a problem where the user faces is learning a fixed constraint in which the systems null-space controller $\bpi$ varies.
For learning a \textit{linear constraint} problem, constant $\bA$ is used and \texttt{\textit{ccl\_learn\_nhat}} is implemented. For learning a \textit{parabolic constraint}, a state-dependent constraint $\bA(\bx)$ of the form $\bA(\bx)=[-2ax,1]$ is used. For this, \texttt{\textit{ccl\_learn\_alpha}} is implemented. Finally, \texttt{\textit{nPPE}} and \texttt{\textit{nPOE}} are used to evaluate the learning performance.



\begin{table}[]
\centering
\caption{Sample code snippet for learning null-space component from Matlab}
\label{t:code_learn_ns}
\scalebox{.67}{%
\begin{tabular}{l}
\hline
\texttt{Data = ccl\_data\_gen(settings) ;}                                                       \\
\texttt{model.c = ccl\_math\_gc (X, model.dim\_basis) ;}                                        \\
\texttt{model.s = mean(mean(sqrt(ccl\_math\_distances(model.c, model.c))))\textasciicircum 2 ;} \\
\texttt{model.phi = @(x)ccl\_basis\_rbf ( x, model.c, model.s);}                                \\
\texttt{model = ccl\_learnv\_ncl (X, U, model) ;}                                               \\
\texttt{f\_ncl = @(x) ccl\_learnv\_pred\_ncl ( model, x ) ;}                                    \\
\texttt{NSp = f\_ncl (X) ;}                                                                     \\
\texttt{NUPE = ccl\_error\_nupe(Un, Unp) ;}                                                     \\
\texttt{NPE = ccl\_error\_npe (Un, Unp) ;}                                                      \\ \hline
\end{tabular}
}
\end{table}


\begin{table}[]
\centering
\caption{To do: Sample code snippet for learning state independent and dependent constraint from matlab}
\label{t:code_learn_A}
\scalebox{.72}{%
\begin{tabular}{l}
\hline
\texttt{Data = ccl\_data\_gen(settings) ;}                                       \\
\texttt{model = ccl\_learn\_nhat (Un) ; \% for learning $\bA$}                  \\
\texttt{model = ccl\_learn\_alpha (Un, X, settings) ; \% for learning $\bA(\bx)$} \\
\texttt{nPPE = ccl\_error\_ppe(U, model.P, Pi) ;}                                \\
\texttt{nPOE = ccl\_error\_poe(U, model.P, Pi) ;}                                \\ \hline
\end{tabular}
}
\end{table}

\begin{table}[t]
\centering
\caption{Sample code snippet for learning null-space control policy from Matlab}
\label{t:code_learn_pi}
\scalebox{.67}{%
\begin{tabular}{l}
\hline
\texttt{Data = ccl\_data\_gen(settings);}                                                       \\
\texttt{model.c = ccl\_math\_gc (X, model.dim\_basis) ;}                                        \\
\texttt{model.s = mean(mean(sqrt(ccl\_math\_distances(model.c, model.c))))\textasciicircum 2 ;} \\
\texttt{model.phi = @(x)ccl\_basis\_rbf ( x, model.c, model.s);}                                \\
\texttt{model = ccl\_learnp\_pi (X, Un, model) ;}                                               \\
\texttt{fp = @(x) ccl\_learnp\_pred\_linear ( model, x ) ;}                                    \\
\texttt{Pi = fp (X) ;}                                                                     \\
\texttt{NUPE = ccl\_error\_nupe(Un, Unp) ;}                                                     \\
\texttt{NCPE = ccl\_error\_ncpe (Un, Unp) ;}                                                      \\ \hline
\end{tabular}
}
\end{table}

\subsubsection{Learning Null Space Policy $\pi$}
This section will give guidance on learning an unconstrained control policy $\bpi$. This applies to the use case where $\bpi$ is consistent but $\bA$ is varying. A sample Matlab script is provided in \tref{code_learn_pi} and explained as the following. For learning, 10 \RBF s are used and learned using the same K-means algorithm. \texttt{\textit{ccl\_learnp\_pi}} is then implemented for training the model. Finally, \texttt{\textit{NUPE}} and \texttt{\textit{NCPE}} are calculated to evaluate the model's performance. In the library, a locally weighted policy learning method is also implemented in both Matlab and C. For details please refer to the documentation.

Other examples such as a \texttt{2-link arm} and \texttt{wiping} examples are also implemented which follow a similar procedure to the toy example but in a higher dimension to aid novice users wanting to take advantage of the \textit{\CCL} to learn the kinematic redundancy of a robot. Moreover, users can easily adapt these learning methods, where the provided demonstrations such as \texttt{wiping} and \texttt{real data} are setup to use the 7 \DoF  KUKA LWR3 and Trakstar 3D electromagnetic tracking sensor, respectively, are reconfigured to suit their own systems and requirements.





\section{Conclusion}\label{s:conclusion}
\noindent This paper introduces the \CCL\ library, an open-source collection of software tools for learning different components of constrained movements and behaviours. The implementations of the key functions have been explained throughout the paper, and interchangeable and expandable examples have been demonstrated using both simulated and real world data for their usage. For the first time, the library brings together a diverse collection of algorithms developed over the past ten years into a unified framework with a common interface for software developers in the robotics community. In the future work, Matlab and python wrappers will be released by taking advantage of the fast computation routine implemented in C.
\bibliographystyle{paper}
\bibliography{bib/abbreviations,bib/bibliography}
\end{document}